\documentclass[review]{elsarticle}

\usepackage{lineno,hyperref}
\usepackage{amsmath}
\usepackage{times}
\usepackage{epsfig}
\usepackage{graphicx}
\usepackage{amsmath}
\usepackage{amssymb}

\usepackage{caption}
\usepackage{algorithm}
\usepackage{algorithmic}
\usepackage{float}
\usepackage{amsmath}
\usepackage{mathrsfs}

\journal{Journal of \LaTeX\ Templates}

\begin{document}

\begin{frontmatter}

\title{Transfer feature generating networks with semantic classes structure for zero-shot learning}

\author[mymainaddress]{Guangfeng Lin}\corref{mycorrespondingauthor}
\cortext[mycorrespondingauthor]{Corresponding author}
\ead{lgf78103@xaut.edu.cn}

\author[mymainaddress]{Wanjun Chen}
\author[mymainaddress]{Kaiyang Liao}
\author[mymainaddress]{Xiaobing Kang}
\author[mymainaddress]{Caixia Fan}


\address[mymainaddress]{Information Science Department, Xi'an University of Technology,\\
 5 South Jinhua Road, Xi'an, Shaanxi Province 710048, PR China}


\begin{abstract}
Feature generating networks face to the most important question, which is the fitting difference (inconsistence) of the distribution between the generated feature and the real data. This inconsistence further influence the performance of the networks model, because training samples from seen classes is disjointed with testing samples from unseen classes in zero-shot learning (ZSL). In generalization zero-shot learning (GZSL), testing samples come from not only seen classes but also unseen classes for closer to the practical situation. Therefore, most of feature generating networks difficultly obtain satisfactory performance for the challenging GZSL by adversarial learning the distribution of semantic classes. To alleviate the negative influence of this inconsistence for ZSL and GZSL, transfer feature generating networks with semantic classes structure (TFGNSCS) is proposed to construct networks model for improving the performance of ZSL and GZSL. TFGNSCS can not only consider the semantic structure relationship between seen and unseen classes, but also learn the difference of generating features by transferring classification model information from seen to unseen classes in networks. The proposed method can integrate the transfer loss, the classification loss and the Wasserstein distance loss to generate enough CNN features, on which softmax classifiers are trained for ZSL and GZSL. Experiments demonstrate that the performance of TFGNSCS outperforms that of the state of the arts on four challenging datasets, which are CUB,FLO,SUN, AWA in GZSL.
\end{abstract}


\end{frontmatter}

\section{Introduction}
\begin{figure*}[ht]
  \begin{center}
\includegraphics[width=0.7\linewidth]{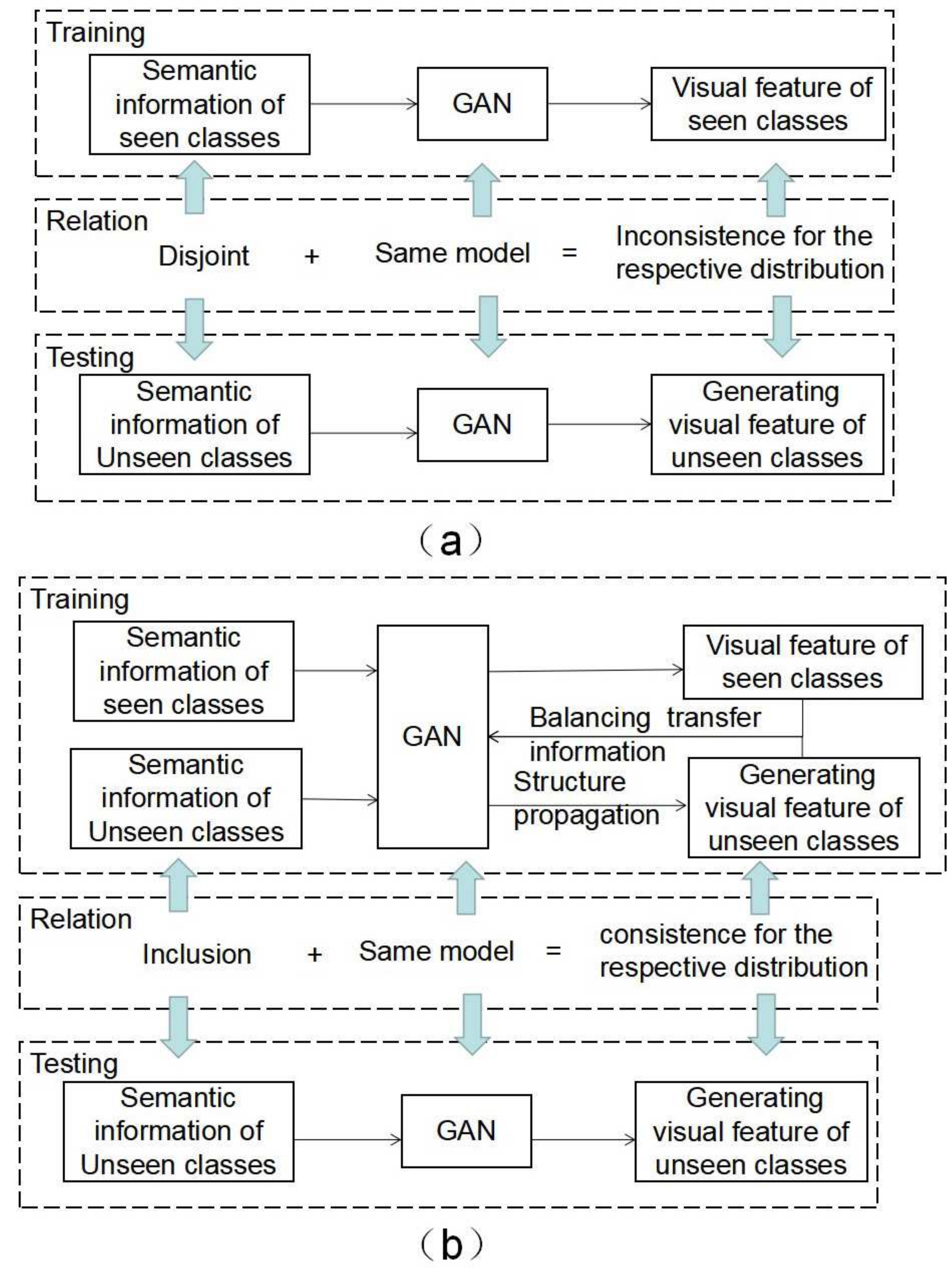}
\end{center}
\vspace{-0.2in}
 \caption{Comparison between generative feature network method in (a) (for example CLSWGAN\cite{YXian2018}) and the proposed method (TFGNSCS) in (b). GAN means generative adversarial network.}
  \label{TFGNSCS1}
 \end{figure*}
Based on large amounts of labeled data training, deep learning can capture the various patterns of data for large-scale recognition problem. However, in many practical application, we usually lack annotated data, which needs lots of time-consume to manually annotate. Therefore, data generation \cite{Goodfellow2014Generative} \cite{8411144} \cite{gulrajani2017improved} \cite{8066319} \cite{YXian2018} with labels has become an important method for obtaining enough annotated data. Generative adversarial networks (GAN) \cite{Goodfellow2014Generative} can synthesize the approximate images on object classes\cite{8411144}\cite{8066319}, but can not generate sufficiently discriminative images or features without classification information. Especially, because training samples from seen classes is disjointed with testing samples from unseen classes in ZSL or GZSL, generative features for different classes don't accurately match with the respective distribution in GAN. In other words, there is some data shift between generative features for unseen classes and their real distribution, since generate networks model is often trained by the samples of seen classes. Existing CLSWGAN\cite{YXian2018} considers the classification loss of seen classes for improving the performance of ZSL or GZSL. However, the classification loss of unseen classes is also important for ZSL or GZSL. Therefore, our motivation is how to transfer classification information from seen to unseen classes to construct the classification loss of unseen classes (this loss is called transfer loss) for learning generate networks model.

ZSL\cite{Lampert2014}\cite{Fu2015Transductive}\cite{Changpinyo2016}\cite{Kodirov2017}\cite{LINGF2018} is an arguable problem about the extreme condition of few samples. Some classes (seen classes) have visual samples , while others (unseen classes) have no visual samples during training in ZSL. In this work, we focus on the transferability of generative adversarial model, and expect to use transfer information to process the generating feature inconsistence of the unbalance learning (In Fig \ref{TFGNSCS1}, we explain this point, which means learning model from seen classes to generate features for unseen classes)between seen and unseen classes for constructing learning model in ZSL and GZSL. Generative features for unseen classes by transfer generative adversarial model is used for the traditional supervised learning to solve ZSL and GZSL. Therefore, the main contribution in our paper is the proposed TFGNSCS based on existing CLSWGAN\cite{YXian2018} to find the importance of transfer information for processing the unbalanced learning between seen and unseen classes. Especially, we look into the influence of the different transfer loss for generating features in ZSL or GZSL. We mainly discuss two transfer losses. One (the detail is defined by equation (\ref{loss3}) in section \ref{FG}) is the consideration of the structure relationship between seen and unseen classes by structure propagation (the details in section \ref{spr}), and the other (the detail is defined by equation (\ref{loss4}) in section \ref{FG})is balancing the difference of generating features between seen and unseen classes by discriminator information. In this motivation, we proposed a novel generative feature GAN method-namely TFGNSCS that is learned with a novel transfer loss improving over existing GAN-models for generating features.

Our contributions have three point as follows.(a) We present a novel adversarial generative model TFGNSCS that synthesizes CNN features of classes by optimizing and balancing the related losses, which are the transfer loss,the classification loss and the Wasserstein distance loss. (b) In four challenging datasets with the different size or granularity, the proposed TFGNSCS outperforms the state of the arts in GZSL setting. (c) Our model is generalized to different transfer information ways for evaluating the performance of generative features model.

\section{Related Works}
The related works of the proposed method involve generative adversarial networks(GAN), structure propagation, zero-shot learning (ZSL) and generalization gero-shot lLearning (GZSL).

\subsection{Generative Adversarial Networks}
GAN\cite{Goodfellow2014Generative} can initially learn a generative model to follow an arbitrary distribution by a discriminative model adjustment, for example images distribution fitting. In terms of GAN theory development, this process involves three aspects. The first aspect is GAN training improvement by additional information, such as deep convolution neural network in DCGAN \cite{radford2015unsupervised}, the style and structure networks in improved DCGAN \cite{wang2016generative}, and the mutual information between the latent variables and the generator distribution in InfoGAN \cite{chen2016infogan}. The second aspect is conditional GAN by feeding the related information into networks, for instance class label \cite{mirza2014conditional} and sentence descriptions \cite{8411144}. The third aspect is about stability GAN training by the relevance constraint \cite{karras2017progressive}, which can be Wasserstein distance \cite{arjovsky2017wasserstein} or Lipschitz constraint \cite{gulrajani2017improved}. Recently, CLSWGAN \cite{YXian2018} can utilize WGAN idea with classification loss for generating image feature, and demonstrate the promising results in ZSL and GZSL.

In this paper, we intuitively find that feature generated by the state of art CLSWGAN \cite{YXian2018}(this model only is trained by the samples of seen classes) are not enough fitting the distribution of unseen classes for learning a classifier. Hence, we present a novel GAN framework to synthesize CNN features to learn a discriminative classifier for ZSL and GZSL. Integrating the promising CLSWGAN \cite{YXian2018} loss and transfer loss which transmits the information of the different classes to generate the discriminative feature, our proposed GAN framework outperforms CLSWGAN \cite{YXian2018} owing to the regularizer of transfer loss.

\subsection{Structure Propagation}
\label{spr}
To best of our knowledge, structure propagation is firstly proposed for image completion  as a global optimization problem by enforcing structure and consistency constraints \cite{sun2005image}. Structure can be defined as the graph structure among data samples and plays a very important role for visual information discrimination. In recent works, there are two kinds of the impressive methods. One is structure information propagation in label space, such as dynamic structure fusion and label propagation to refining the relation of objects for semi-supervised multi-modality classification \cite{Lin2017Dynamic} and information propagation mechanism from the semantic label space, which can be applied to model the interdependencies between seen and unseen class labels \cite{lee2018multi}. The other is structure information propagation between seen and unseen classes, for instant structure fusion and propagation to update the relevance of multi-semantic classes by the iteration computation for ZSL \cite{Lin2018structure} \cite{LINGF2018}, structure propagation constraining the encoder-decoder mechanism of the bidirectional projection for ZSL \cite{lin2018class}, and absorbing Markov chain process propagation constructing semantic class prototype graph for ZSL \cite{fu2018zero}. Although those papers have shown the information transferability of structure propagation, structure propagation is not used for generating feature in adversarial mechanism to balance the difference between seen and unseen classes.

In this paper, we expect to construct a novel GAN framework by transfer loss adding into CLSWGAN \cite{YXian2018}. Transfer loss includes two parts. One is balancing transfer information by generative model iteration, and the other is structure propagation for accurately computing classification loss of all classes. In contrast, CLSWGAN \cite{YXian2018} only calculates classification loss of seen classes. Therefore, structure propagation can further extend CLSWGAN \cite{YXian2018} model to relieve the generating feature inconsistence of seen classes training model for following the distribution of unseen classes.

\subsection{Zero-shot Learning}
In ZSL, seen classes of model learning in training and unseen classes of model evaluation in testing are disjoint \cite{larochelle2008zero}. According to the utilization of deep network framework in ZSL, ZSL methods can be divided into two categories to bridge the gap between seen and unseenclasses. One involves non-deep network framework for ZSL, such as semantic attribute classifiers learning\cite{Rohrbach2010What} \cite{Rohrbach2011Evaluating} \cite{Lampert2014}, seen class proportions combining unseen class \cite{zhang2015zero} \cite{7781018} \cite{norouzi2013zero} \cite{Changpinyo2016}, and the learning compatibility between images and classes \cite{Akata2016Label} \cite{7298911} \cite{Socher2013Zero} \cite{romera2015embarrassingly} \cite{7780384} \cite{Kodirov2017}. The other category utilizes deep network framework for ZSL, for instance DeViSE model\cite{Frome2013DeViSE},latent discriminative feature learning (LDF) model \cite{li2018discriminative}  and quasi-fully supervised learning (QFSL) model by deep network optimizing the visual or semantic model, synthesizing example \cite{verma2018generalized} or preserving semantic relation \cite{annadani2018preserving} by autoencoder architecture, multi-label zero-shot learning (ML-ZSL) \cite{lee2018multi} or graph convolutional network for zero-shot learning \cite{wang2018zero} with the benefit of knowledge graph,and semantics-preserving adversarial embedding network (SP-AEN) \cite{chen2018zero} or feature generating network \cite{zhu2018generative} \cite{YXian2018} based on generation adversarial mechanism.

In summary, generation adversarial frameworks demonstrate the promising results, and especially visual feature generation outperforms image generation based on the same adversarial frameworks for ZSL. However, these frameworks rarely consider transfer information based on structure propagation for finding the more discriminative feature in ZSL. Therefore, for considering the transfer loss constrains,we expect to construct a novel generation adversarial frameworks to capture discriminative information for tackling ZSL or GZSL.

\section{Transfer Feature Generating Networks}
Existing ZSL methods only use the information of seen classes (label data, image feature data and semantic data) during training, and predict the label of unseen classes by the potential relation of semantic space (In a complete semantic space, semantic concepts have an uniform description, on which their distribution relation can be captured.). The main idea of the proposed model transfers the information of seen classes into synthesized feature of unseen classes by structure propagation, and iteratively constructs the learned model based on the real information of seen classes and the generative information of unseen classes. Therefore, the key point of the proposed method not only draws support from semantic embedding vector but also models transfer relation (structure propagation) to generate CNN features without any images of the class. The transfer relation can alleviate the inconsistence of the generation distribution on all categories. Because we can use synthesized CNN features as samples of unseen classes, ZSL can be converted into supervised learning. We can train Softmax classifier for recognizing unseen classes.

We define the following notation for describing ZSL or GZSL. $S= \{(x_{s}, y_{s}, c(y_{s}))|x_{s} \in\mathbb{X},y_{s}\in \mathbb{Y}_{s}, c(y_{s})\in \mathbb{C}\}$ is training set, which includes seen classes. $x_{s}\in \mathbb{R}^{d_{x}}$ is the CNN feature with $d_{x}$ dimension in $\mathbb{X}$ feature sets, $y_{s}$ stands for the class label of $x_{s}$ in $\mathbb{Y}_{s}=\{y_{s}|s=1,2,...,K\}$, and $c(y_{s})\in \mathbb{R}^{d_{c}}$ denotes the class $y_{s}$ of the semantic embedding, which represents the class vector of semantic description(such as attributes), in $\mathbb{C}$ semantic embedding sets.In addition, the available information of unseen classes in training is $U=\{(y_{u},c(y_{u}))|y_{u}\in \mathbb{Y}_{u},c(y_{u})\in \mathbb{C}\}$ where $y_{u}$,$c(y_{u})$ and $\mathbb{Y}_{u}=\{y_{u}|u=1,2,...,M\}$ is respectively the class label, the class embedding and the class label set in unseen classes without image and feature. Therefore, the purpose of ZSL is to learn a projection $f: \mathbb{X}\times \mathbb{C}\rightarrow\mathbb{Y}_{s}$ for discriminating images of unseen classes belonging to which one in $\mathbb{Y}_{u}$, while the task of GZSL learns the same projection  for recognizing seen and unseen classes of images being which one of $\mathbb{Y}_{s}\bigcup \mathbb{Y}_{u}$, where $\mathbb{Y}_{s}\bigcap\mathbb{Y}_{u}=\varnothing$.

\subsection{Feature generation}
\label{FG}
In this section, we discuss the feature generation model CLSWGAN \cite{YXian2018} based on GAN framework as the proposed model basis. The main idea of GAN is playing a game between a generative network $G$ and a discriminative $D$ to optimize data generation following the specialization distribution. In CLSWGAN, $D$ need identify as much as possible real feature from generated feature, while $G$ need trick the discriminator by generated feature that has deviation compared with the real feature. Compared CLSWGAN with GAN, the differences are the addition of the classification loss and the metric method change of the Wasserstein distance loss. According to the inspiration of conditional GAN \cite{mirza2014conditional}\cite{8411144}, we expect to extend CLSWGAN to the proposed TFGNSCS with a conditional transfer transformation to both $G$ and $D$. In the following we describe the details of TFGNSCS, the novelty of which is that we introduce transfer information into the conditional GAN to generate the more discriminative features for ZSL or GZSL. It is worth noting that the proposed TFGNSCS not only can synthesize the good fidelity features of unseen classes in $S$, but also can refine the performance of the model by the generated features of unseen classes.

We can extend CLSWGAN \cite{YXian2018} to the proposed TFGNSCS by transferring the probability model of the generated features from seen to unseen classes for the adversarial training between the generator and the discriminator. The loss has four parts. The first part is constructed  based on the improved WGAN\cite{gulrajani2017improved} and conditional WGAN\cite{YXian2018} with the class embedding $c(y_{s})$.
\begin{align}
\label{loss1}
\begin{aligned}
&L_{WGAN}=E[D(x_{s},c(y_{s}))]-E[D(\tilde{x}_{s},c(y_{s}))]-\lambda E[(\|\triangledown_{\hat{x}}D(\hat{x},c(y_{s}))\|_{2}-1)^2],
 \end{aligned}
\end{align}
where $\tilde{x}_{s}=G(z,c(y_{s}))$ is the generative feature of seen classes, $z\in \mathbb{Z}\subset \mathbb{R}^{d_{z}}$ is random Gaussian noise, $c(y_{s})\in \mathbb{C}$ is class embedding of seen classes, $\hat{x}=\alpha x_{s}+(1-\alpha)\tilde{x}$ with $\alpha \thicksim U(0,1)$, and $\lambda$ is the trade-off coefficient. In the loss $L_{WGAN}$, the first two terms compute the Wasserstein distance, and the third term constrains the gradient of $D$ to become unit norm following the straight line between pairs of real and generated point\cite{YXian2018}.

The second part of the loss is expected to generate CNN feature for adapting a discriminative classifier. In other word, the construction of the classifier can constrain the feature generation of $G$ for balancing their relationship. Therefore, we can maximize the probability of the generated feature $\tilde{x}_{s}$ in the classifier trained by the real feature $x_{s}$, and further minimize the classification loss, which is defined by the negative log likelihood.
\begin{align}
\label{loss2}
\begin{aligned}
&L_{CLS}=-E_{\tilde{x}_{s}\thicksim p_{\tilde{x}_{s}}}[\log P(y_{s}|\tilde{x}_{s};\theta)],
 \end{aligned}
\end{align}
where $\tilde{x}_{s}=G(z,c(y_{s}))$, $y_{s}$ is the class label of $\tilde{x}_{s}$ in seen classes, $P(y_{s}|\tilde{x}_{s};\theta)$ is the probability of $\tilde{x}_{s}$ with the class label $y_{s}$. The probability can be modeled by the $\theta$ parameterizing classification methods, for example the linear softmax classifier or support vector machine. These classification methods can be learned by the real feature and the class label pairs in seen classes.

The first two parts of the loss are the main ideas of CLSWGAN \cite{YXian2018}. We propose the novel transfer loss( includes the third part and the forth part of the loss) based on CLSWGAN \cite{YXian2018} for considering transfer information. Therefore, we expect that the third part of the loss can process the classification loss of the generated feature $\tilde{x}_{u}$ in unseen classes. However, we can not construct a classifier trained by the real feature $x_{u}$ that is lost in ZSL or GZSL. Because semantic information is complete in the concept of all classes, we can draw support from the relationship between seen and unseen classes in semantic embedding to transfer the classifier model $P(\theta)$. we define that $T(P(\theta))$ is transfer transformation function (It is explained in Section \ref{transfer}), and the third part of the loss is the model transfer loss $L_{TRA1}$ that is
\begin{align}
\label{loss3}
\begin{aligned}
&L_{TRA1}=-E_{\tilde{x}_{u}\thicksim p_{\tilde{x}_{u}}}[\log Q(y_{u}|\tilde{x}_{u};\theta)],
 \end{aligned}
\end{align}
where $\tilde{x}_{u}=G(z,c(y_{u}))$, $c(y_{u})$ is class embedding of semantic description in unseen classes, $Q(y_{u}|\tilde{x}_{u};\theta)$ is the probability of $\tilde{x}_{u}$ with the class label $y_{u}$ based on classifier model $Q(\theta)=T(P(\theta))$. The probability of unseen classes is parameterized by $\theta$. $L_{TRA1}$ is the negative log-likelihood of the transformed model $T(P(\theta))$.In $L_{TRA1}$, we constrain the discrimination of generative features in unseen classes by transferring the classification model of seen classes, and further alleviate the generative feature inconsistence between seen and unseen classes.

The forth part of the loss is constructed based on the discriminator $D$. We expect to capture the identification information  of unseen classes in the discriminator $D$. Therefore, the loss $L_{TRA2}$ can be defined as
\begin{align}
\label{loss4}
\begin{aligned}
&L_{TRA2}=-E[D(\tilde{x}_{u},c(y_{u}))],
 \end{aligned}
\end{align}
where $\tilde{x}_{u}=G(z,c(y_{u}))$,$c(y_{u})$ is class embedding of semantic description in unseen classes. In the loss $L_{TRA2}$, we consider the information of unseen classes for balancing the bias of the discriminator $D$ in seen classes. Therefore, the total loss can be built by the above four parts losses, and full objective can be reformulated as following,
\begin{align}
\label{loss5}
\begin{aligned}
&\min_{G}\max_{D}L=&\min_{G}\max_{D}L_{WGAN}+\beta L_{CLS}+ \gamma L_{TRA1}+ \eta L_{TRA2},
 \end{aligned}
\end{align}

\begin{figure*}[ht]
  \begin{center}
\includegraphics[width=0.9\linewidth]{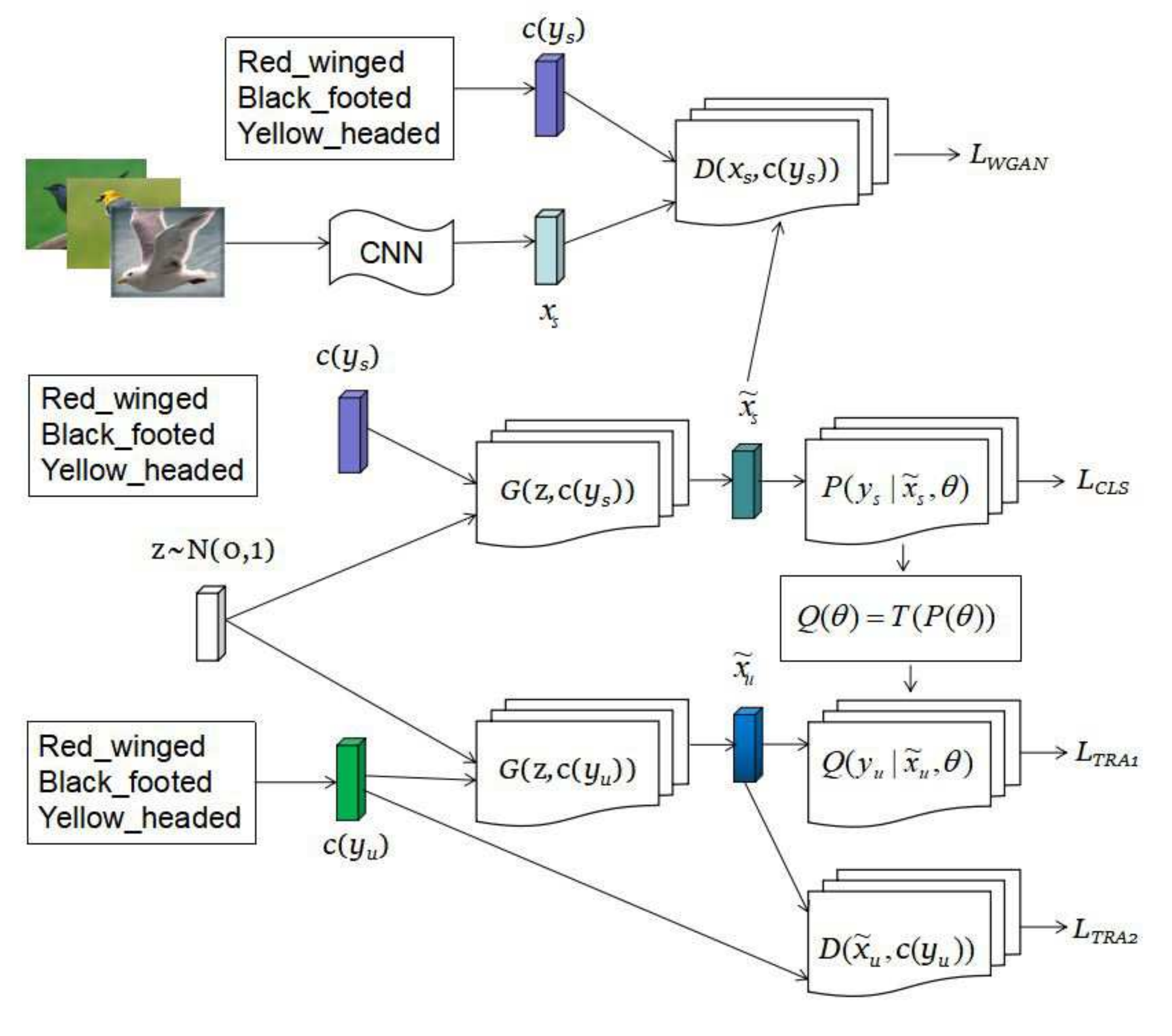}
\end{center}
\vspace{-0.2in}
 \caption{The proposed TFGNSCS can minimize the transfer loss over the unseen classes classification loss $L_{TRA1}$ and the generated features discrimination loss $L_{TRA2}$.$T(P(\theta))$ is the transfer transformation function,$\tilde{x}_{s}$ and $\tilde{x}_{u}$ are respectively the generation feature of seen and unseen classes.}
  \label{fig1}
 \end{figure*}

\begin{figure*}[ht]
  \begin{center}
\includegraphics[width=0.7\linewidth]{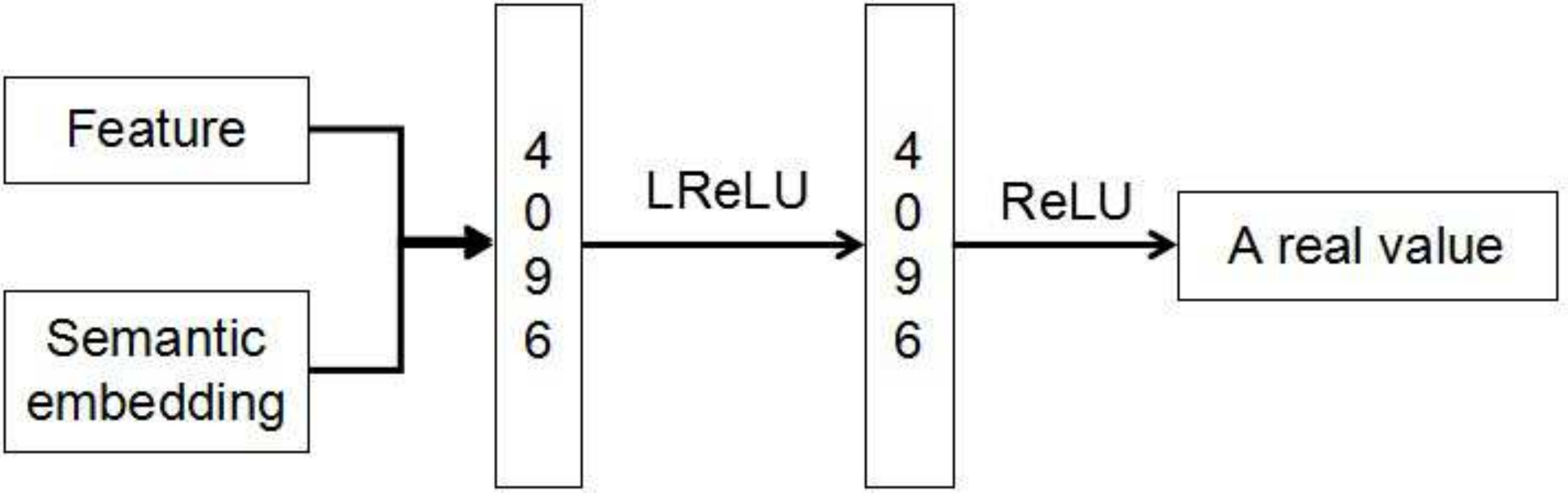}
\end{center}
\vspace{-0.2in}
 \caption{The architecture of discriminator $D$. The output of discriminator is a real value that represents the matching degree between feature and semantic embedding in the input of discriminator.}
  \label{fig2}
 \end{figure*}

 \begin{figure*}[ht]
  \begin{center}
\includegraphics[width=0.7\linewidth]{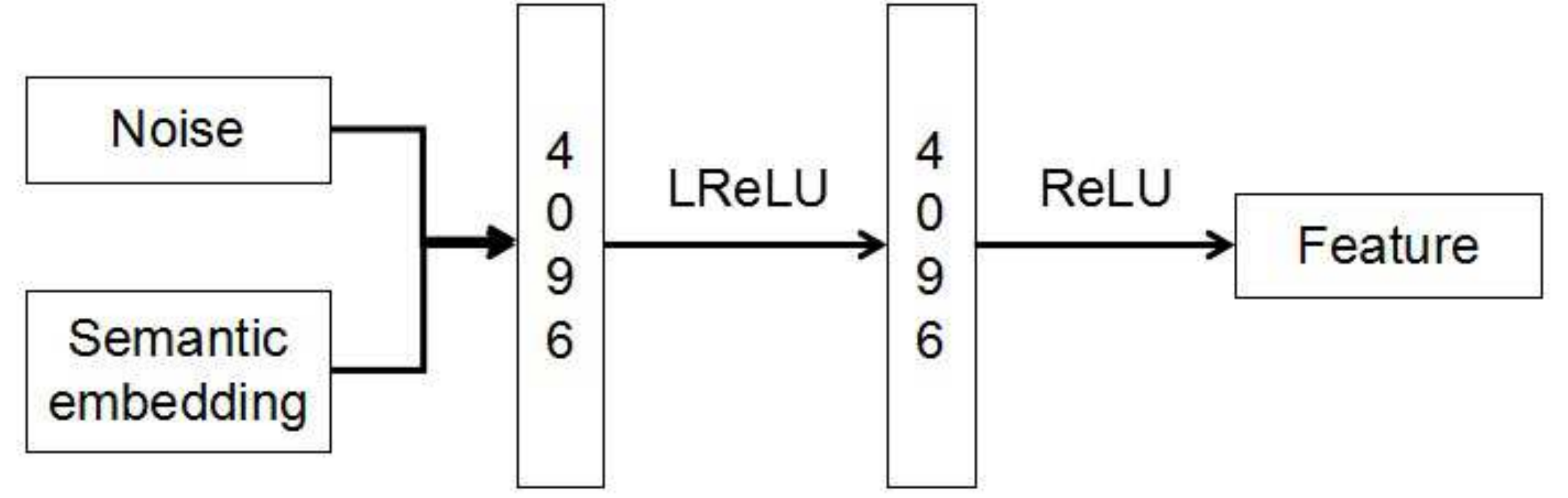}
\end{center}
\vspace{-0.2in}
 \caption{The architecture of generator $G$.}
  \label{fig3}
 \end{figure*}

\subsection{Transferring and Classification}
\label{transfer}
In the proposed model, we need obtain three classifier model to generate the discriminative feature for classifying unseen classes.

The first classifier model $P(\theta)$ can be trained by samples of seen classes for describing $L_{CLS}$ to improve the classification performance of the generative feature of seen classes. We can learn model parameter $\theta$ by following equation that is
\begin{align}
\label{model1}
\begin{aligned}
&\min_{\theta} -\frac{1}{N_{\mathbb{T}}}\sum_{(x_{s},y_{s})\in \mathbb{T}}\log P(y_{s}|x_{s};\theta),
\end{aligned}
\end{align}
where $P(y_{s}|x_{s};\theta)=\frac{\exp(\theta^{T}_{s}x_{s})}{\sum_{s}\exp({\theta^{T}_{s}x_{s}})}$, $P(\theta)=[P(y_{1}|x_{1};\theta),P(y_{2}|x_{2};\theta),...,P(y_{K}|x_{K};\theta)]$, $\theta_{s}$ is a column of $\theta\in \mathbb{R}^{d_{x}\times K}$ that is transformation matrix corresponding to image feature $x$ to $K$ classes probabilities mapping, $\mathbb{T}=S$, and $N_{\mathbb{T}}$ is the number of $\mathbb{T}$.

The second classifier model $Q(\theta)$ can not be learned by unseen classes of samples, which is lost in ZSL and GZSL. Therefore, we construct $Q(\theta)$ by $T(P(\theta))$ for representing $L_{TRA1}$ to enhance the discrimination of the synthesized feature of unseen classes. Given $P(\theta)=A_{s}W_{ss}$, $P(\theta)$ can be decomposed into the sharing part $A_{s}$ and the unique part $W_{ss}$ in probability pattern of seen classes.$Q(\theta)=A_{u}W_{uu}$ means that $Q(\theta)$ can be divided into the sharing part $A_{u}$ and the unique part $W_{uu}$ in probability pattern of unseen classes. $A_{u}=A_{s}W_{su}$ indicates the relationship of the sharing parts between seen and unseen classes. Therefore, transfer transformation function $T(P(\theta))$ can be deduced as
\begin{align}
\label{model2}
\begin{aligned}
&Q(\theta)=P(\theta)W_{ss}^{-1}W_{su}W_{uu},
\end{aligned}
\end{align}
where$W_{ss}\in \mathbb{R}^{K\times K}$ is the similarity matrix among seen classes,$W_{uu}\in \mathbb{R}^{M\times M}$ is the similarity matrix among unseen classes,$W_{su}\in \mathbb{R}^{K\times M}$ is the similarity matrix between seen and unseen classes based on semantic class prototype graph(In this graph, each class is corresponding to one node, which can be represented by the semantic embedding of each class. The weight between nodes can describe the similarity between classes). These similarity matrices (structure representation) can be measured by cosine distance $d(a,b)$ between the semantic embedding $a$ and $b$ in any two classes.
\begin{align}
\label{distance1}
\begin{aligned}
&d(a,b)=\frac{a\cdot b}{\|a\|\cdot\|b\|},
\end{aligned}
\end{align}

\begin{align}
\label{distance2}
  W_{ss}(i,j)=\left\{
   \begin{aligned}
      &d(c(y_{s})_{i},c(y_{s})_{j}),~c(y_{s})_{i}\in N_{c(y_{s})_{j}}\\
      &0~~~~\mbox{else,}
   \end{aligned}
  \right.
\end{align}
where $W_{ss}(i,j)$ is a element of $W_{ss}$, $i=\{1,2,...,K\}$, $j=\{1,2,...,K\}$, $c(y_{s})_{i}$ or $c(y_{s})_{j}$ denotes the semantic embedding of any class $i$ or $j$ in seen classes,$N_{c(y_{s})_{j}}$ is the neighborhood of $c(y_{s})_{j}$. For selecting the related semantic embedding, the neighborhood number often is set as $5$.
\begin{align}
\label{distance22}
  W_{uu}(i,j)=\left\{
   \begin{aligned}
      &d(c(y_{u})_{i},c(y_{u})_{j}),~c(y_{u})_{i}\in N_{c(y_{u})_{j}}\\
      &0~~~~\mbox{else,}
   \end{aligned}
  \right.
\end{align}
where $W_{uu}(i,j)$ is a element of $W_{uu}$, $i=\{1,2,...,M\}$, $j=\{1,2,...,M\}$, $c(y_{u})_{i}$ or $c(y_{u})_{j}$ denotes the semantic embedding of any class $i$ or $j$ in unseen classes,$N_{c(y_{u})_{j}}$ is the neighborhood of $c(y_{u})_{j}$.For selecting the related semantic embedding, the neighborhood number often is set as $5$.
\begin{align}
\label{distance3}
  W_{su}(i,j)=\left\{
   \begin{aligned}
      &d(c(y_{s})_{i},c(y_{u})_{j}),~c(y_{s})_{i}\in N_{c(y_{u})_{j}}\\
      &0~~~~\mbox{else,}
   \end{aligned}
  \right.
\end{align}
where $W_{su}(i,j)$ is a element of $W_{su}$, $i=\{1,2,...,K\}$, $j=\{1,2,...,M\}$, $c(y_{s})_{i}$ stands for the semantic embedding of any class $i$ in seen classes, while $c(y_{u})_{j}$ is the semantic embedding of any class $j$ in unseen classes,,$N_{c(y_{s})_{j}}$ is the neighborhood of $c(y_{s})_{j}$.For selecting the related semantic embedding, the neighborhood number often is set as $5$.

The third classifier model is be constructed based on the real feature of seen classes $s$ and the synthesized feature of unseen classes $u$ for transforming ZSL to supervised learning.We can learn model parameter $\phi$ by following equation that is
\begin{align}
\label{model3}
\begin{aligned}
&\min_{\phi} -\frac{1}{N_{\mathbb{T}}}\sum_{(x,y)\in \mathbb{T}}\log P(y|x;\phi),
\end{aligned}
\end{align}
where $P(y|x;\phi)=\frac{\exp(\phi^{T}_{i}x)}{\sum_{i}\exp({\phi^{T}_{i}x})}$,
$\phi_{i}$ is a column of $\phi\in \mathbb{R}^{d_{x}\times N}$ that is regarded as the weight matrix to project the feature $x$ to $N$ categories in a fully connected layer of deep network. In ZSL, $\mathbb{T}=\tilde{U}$, while $\mathbb{T}=S\bigcup \tilde{U}$ in GZSL, $\tilde{U}=\{U, \tilde{x}_{u}\}$.The prediction function $f(x)$ can be defined as
\begin{align}
\label{model4}
\begin{aligned}
&f(x)=\arg \max P(y|x;\phi),
\end{aligned}
\end{align}
where in ZSL, $x\in \{x_{u}\}$,$x_{u}$ is the image feature of unseen classes and $y\in \mathbb{Y}_{u}$, while in GZSL, $x\in \{x_{s},x_{u}\}$ and $y\in (\mathbb{Y}_{s}\bigcup\mathbb{Y}_{u})$.

The pseudo code of the TFGNSCS can be shown in Algorithm \ref{algTFGNSCS}, which has five steps. The first step (line 1) initializes the structure representation of semantic embedding by equation (\ref{distance1}), (\ref{distance2}), (\ref{distance22})and (\ref{distance3}). The second step (line 2) trains the classifier model of seen classes by equation (\ref{model1}) and transfers the classifier model for unseen classes by equation (\ref{model2}).The third step (line 4) updates the discriminator $D$ with equation (\ref{loss5}). The forth step (from line 5) updates the generator $G$ with equation (\ref{loss5}). The fifth step (from line 7) implements the classifier model training and the label estimating of unseen classes by equation (\ref{model3}) and (\ref{model4}).
\begin{algorithm}[ht]
  \caption{The pseudo code of the TFGNSCS algorithm}
 \begin{algorithmic}[1]
 \label{algTFGNSCS}
\renewcommand{\algorithmicrequire}{\textbf{Input:}}
\renewcommand{\algorithmicensure}{\textbf{Output:}}
\renewcommand{\algorithmicreturn}{\textbf{Iteration:}}
   \REQUIRE $S$ and $U$
   \ENSURE $\hat{y}$ (the estimation value of $y_{u}$ for ZSL or the estimation value of $y_{s}$ and $y_{u}$ for GZSL)
   \STATE Computing the semantic embedding of the structure representation $W_{ss}$ and $W_{su}$ by equation (\ref{distance1}), (\ref{distance2}), (\ref{distance22}) and (\ref{distance3})
   \STATE Training and transferring the classifier model by equation (\ref{model1}) and (\ref{model2})
   \FOR {$1<t<T$}
   \STATE Updating the discriminator $D$ by equation (\ref{loss5})
   \STATE Updating the generator $G$ by equation (\ref{loss5})
   \ENDFOR
   \STATE Training the classifier model and estimating the label $\hat{y}$ of classes by equation (\ref{model3}) and (\ref{model4})
  \end{algorithmic}
\end{algorithm}
\section{Experiments}
We firstly explain our experimental configuration, and then we demonstrate (a) the comparison results between the proposed method and the state of the arts for ZSL and GZSL on four challenging datasets, (b)our analysis for the base-line methods based on the different loss combination, (c)our extending experiments on the different transfer method and (d)our parameter analysis for image feature generation.
\subsection{Datasets}
We implement and evaluate the proposed method TFGNSCS for ZSL or GZSL in four challenging datasets, which are Animals with Attributes (AwA)\cite{Lampert2014}, CUB-200-2011 Birds (CUB)\cite{Wah2011The}, SUN Attribute (SUN)\cite{Patterson2012SUN} and Oxford Flower (FLO)\cite{Nilsback2009Automated}. AwA includes 30475 images, 50 categories and 85 attributes, and belongs to the coarse-grained datasets. CUB, SUN and FLO pertain to the fine-grained datasets. CUB contains 200 birds classes with 312 attributes for a grand total of 11788 images. SUN involves 14340 images from 717 scenes with 102 attributes. FLO has 8189 images from 102 flower classes that can be annotated by the visual description \cite{7780382}. Table \ref{table1} shows the statistics of these datasets.

\begin{table*}[!ht]
\small
\renewcommand{\arraystretch}{1.0}
\caption{Datasets statistics involve semantic embedding $\mathbb{C}$(att/dimension for attribute per class or stc/dimension for sentences), number of classes in training classes ($\mathbb{Y}_{s}$ includes training + validation) and test classes ($\mathbb{Y}_{u})$, visual feature $\mathbb{X}$ in experiments. }
\label{table1}
\begin{center}
\newcommand{\tabincell}[2]{\begin{tabular}{@{}#1@{}}#2\end{tabular}}
\begin{tabular}{lp{1.5cm}p{1.5cm}p{1.5cm}p{1cm}p{2cm}p{1cm}}
\hline
\bfseries Datasets & \bfseries $\mathbb{C}$ & \bfseries $|\mathbb{Y}_{s}|+|\mathbb{Y}_{u}|$& \bfseries $|\mathbb{Y}_{s}|$ & \bfseries $|\mathbb{Y}_{u}|$ &\bfseries $\mathbb{X}$ \\
\hline \hline
AwA  & att/85 &$50$ &$27+13$ & $10$ & \tabincell{l}{Deep feature/2048 \\ by ResNet\cite{7780459}}\\
\hline
CUB  & att/312 &$200$ &$100+50$ &$50$ & \tabincell{l}{Deep feature/2048 \\ by ResNet\cite{7780459}}\\
\hline
SUN  & att/102 &$717$& $580+65$ &$72$& \tabincell{l}{Deep feature/2048 \\ by ResNet\cite{7780459}}\\
\hline
FLO  & stc/1024\cite{7780382} &$102$& $62+20$ &$20$ & \tabincell{l}{Deep feature/2048 \\ by ResNet\cite{7780459}}\\
\hline
\end{tabular}
\end{center}
\end{table*}

\subsection{Visual and Semantic Feature}
\label{feature}
ZSL can recognize the visual samples of unseen classes by the completed semantic relation. Visual feature and semantic class embedding should first be extracted or described. Deep network shows the outstanding performance for extracting the discriminative feature from visual or semantic information. Therefore, we use the same description in \cite{YXian2018}. We can represent the entire image as the 2048 dimension visual feature from the top layer of the pre-trained 101-layer ResNet\cite{7780459} based on ImageNet 1K without image pre-processing, network fine-tuned and data augmentation. We can utilize pre-annotated attributes as semantic class embedding, such as AwA with 85 dimension vector, CUB with 312 dimension vector and SUN with 102 dimension. For FLO without the pre-annotated attributes, we extract 1024 dimension feature based on CNN-RNN of fine-grained visual description\cite{7780382}. In the whole feature extracting process, we obey the ZSL rules that any information of $\mathbb{Y}_{s}$ and $\mathbb{Y}_{u}$ have no crossed set.

\subsection{Classification protocols}
In ZSL, the test image can be corresponding to an unseen class label in $\mathbb{Y}_{u}$, while this image can be related to any one class label in $\mathbb{Y}_{s}\bigcup\mathbb{Y}_{u}$ in GZSL. For evaluating the performance of ZSL or GZSL, we compute the average per-class top-1 accuracy by dividing the each class accuracy sum by the number of classes, for example, $tr$ stands for average per-class top-1 accuracy on seen classes, $ts$ denotes  average per-class top-1 accuracy on unseen classes, and $H=2*tr*ts/(tr+ts)$ represents harmonic mean on $\mathbb{Y}_{s}\bigcup\mathbb{Y}_{u}$ in GZSL. We expect to generate the discriminative feature by the proposed model. For comparing with state of the arts, we preserve the architecture of generator and discriminator in \cite{YXian2018}. These architectures both include multi-layer perception (MLP) with leaky rectified linear unit (LReLU), a single layer with 4096 hidden units and a output rectified linear unit (ReLU) layer for learning top max-pooling units of ResNet-101. The noise $z$ come from Gaussian distribution with a unit variance, and the dimensionality of $z$ is same to that of class embedding. We adopt $\lambda=10$ \cite{gulrajani2017improved}, $\beta=0.01$ \cite{YXian2018},$\gamma=0.01$ and $\eta=1$ for conveniently comparing the different method result. However, all kinds of hyperparameters often can be obtained by cross-validation based on the specific dataset in fact.

\subsection{Comparison with the state-of-the-arts}
\label{State-of-the-arts}
In this section, because generation adversarial architecture and structure constrains are basic ideas for constructing TFGNSCS, we compare the proposed method with five related state-of-the-arts. The first method is dual-verification network (DVN) constructs and verifies the orthogonal projection between features and attributes with a pairwise manner in the respective spaces\cite{ZHANG201943}. The second method is a hybrid model (HM) includes random attribute selection (RAS) and conditional generative adversarial network (cGAN) for adversarial unseen visual feature synthesis\cite{ZHANG201912}. The third method is visual center learning (VCL) can align the projected semantic center and visual cluster center by minimizing the distance between the synthetic and real center in visual feature space\cite{wan019}. The forth method is triple verification network (TVN) can construct a unified optimization of regression and compatibility functions for integrating the complementary losses and the mutual regularization \cite{8464092}. The fifth method is feature generating networks (FGN) can pair a wasserstein GAN with a classification loss to generate sufficiently discriminative CNN features for training softmax classifier in ZSL or GZSL \cite{YXian2018}. One thing to note, the above methods are inductive methods, which do not use test datasets for training the learning model, for strictly following ZSL or GZSL setting.

Tab.\ref{table2} shows the comparison results between the proposed method (TFGNSCS) and five state-of-arts (DVN, HM, VCL, TVN and FGN)for ZSL. The proposed method TFGNSCS has the better result in the various datasets. The performance of TFGNSCS respectively improves $1.7\%$ for AwA, $0.6\%$ for CUB, $2.4\%$ for SUN, and $0.9\%$ for FLO at least.

\begin{table}[!ht]
\small
\renewcommand{\arraystretch}{1.0}
\caption{Comparison of TFGNSCS method with state of the art methods for ZSL with semantic feature and ResNet visual feature.}
\label{table2}
\begin{center}
\newcommand{\tabincell}[2]{\begin{tabular}{@{}#1@{}}#2\end{tabular}}
\begin{tabular}{lp{1cm}p{1cm}p{1cm}p{1cm}}
\hline
\bfseries  &\bfseries ZSL &\bfseries  &\bfseries &\bfseries \\
\cline{2-5}
\bfseries Method &\bfseries AwA &\bfseries CUB &\bfseries SUN &\bfseries FLO  \\
\cline{2-5}
\hline \hline
DVN\cite{ZHANG201943}  & $67.7$   &$57.8$ & $62.4$   &NA   \\
\hline
HM\cite{ZHANG201912}  & $67.4$   &$52.6$ & $61.7$   &NA  \\
\hline
VCL\cite{wan019}  & $61.5$   &$59.6$ & $59.4$   &NA  \\
\hline
TVN\cite{8464092} & $68.8$   &$58.1$ & $60.7$   &NA  \\
\hline
FGN\cite{YXian2018}  & $68.2$   &$57.3$  & $60.8$   &$67.2$ \\
\hline\hline
TFGNSCS & $\textbf{70.5}$   &$\textbf{60.2}$ & $\textbf{64.8}$ & $\textbf{68.1}$   \\
\hline
\end{tabular}
\end{center}
\end{table}

Tab.\ref{table3} demonstrates the comparison results between the proposed method (TFGNSCS) and five state-of-arts (DVN, HM, VCL, TVN and FGN)for GZSL. TFGNSCS outperforms five state-of-arts in all datasets. Harmonic mean $H$ can measure the performance of the different methods in the various datasets. The higher value of $H$ indicates the better result for GZSL. $H$ of TFGNSCS respectively improves $4.9$ for AwA, $3.3$ for CUB, $1.2$ for SUN, and $3.1$ for FLO at least.

\begin{table}[!ht]
\small
\renewcommand{\arraystretch}{1.0}
\caption{Comparison of TFGNSCS method with state of the art methods for GZSL with semantic feature and ResNet visual feature. tr=average per-class Top-1 accuracy (\%) on seen classes, ts=average per-class Top-1 accuracy (\%) on unseen classes, H=harmonic mean for GZSL are reported based on the same data configurations in the different datasets splits. }
\label{table3}
\begin{center}
\newcommand{\tabincell}[2]{\begin{tabular}{@{}#1@{}}#2\end{tabular}}
\begin{tabular}{cp{0.4cm}p{0.4cm}p{0.4cm}|p{0.4cm}p{0.4cm}p{0.4cm}|p{0.4cm}p{0.4cm}p{0.4cm}|p{0.4cm}p{0.4cm}p{0.4cm}}
\hline
\bfseries   &\multicolumn{12}{c}{GZSL} \\ \cline{2-13}
\bfseries&\bfseries AwA &\bfseries &\bfseries &\bfseries CUB &\bfseries &\bfseries &\bfseries SUN &\bfseries &\bfseries &\bfseries FLO &\bfseries &\bfseries \\
\cline{2-13}
\bfseries Method &\bfseries ts &\bfseries tr &\bfseries H &\bfseries ts &\bfseries tr &\bfseries H &\bfseries ts &\bfseries tr &\bfseries H &\bfseries ts &\bfseries tr &\bfseries H\\
\hline \hline
DVN\cite{ZHANG201943}  & $34.9$ & $73.4$ & $48.5$ & $26.2$ & $55.1$ & $35.5$ & $25.3$ & $34.6$ & $29.2$ & NA & NA & NA \\
\hline
HM\cite{ZHANG201912}  & $38.7$ & $74.6$ & $51.0$ & $31.5$ & $40.2$ & $35.3$ & $41.2$ & $26.7$ & $32.4$ & NA & NA & NA\\
\hline
VCL\cite{wan019}  & $21.4$ & $\textbf{89.6}$ & $34.6$ & $15.6$ & $\textbf{86.3}$ & $26.5$  & $10.4$ & $\textbf{63.4}$ & $17.9$ & NA & NA & NA\\
\hline
TVN\cite{8464092} & $27.0$ & $67.9$ & $38.6$ & $26.5$ & $62.3$ & $27.2$ & $22.2$ & $38.3$ & $28.1$ & NA & NA & NA \\
\hline
FGN\cite{YXian2018}  & $57.9$ & $61.4$ & $59.6$  &$43.7 $ & $57.7$ & $49.7$ &$42.6$ & $36.6$ & $39.4$ &$59.0$ & $73.8$ & $65.6$ \\
\hline\hline
TFGNSCS & $\textbf{60.3}$ & $69.5$ & $\textbf{64.5}$ & $\textbf{47.6}$ & $59.7$ & $\textbf{53.0}$ & $\textbf{44.3}$ & $37.5$ & $\textbf{40.6}$ & $\textbf{61.1}$ & $\textbf{78.6}$ & $\textbf{68.7}$ \\
\hline
\end{tabular}
\end{center}
\end{table}

\subsection{Comparison with the base-line methods}
\label{base-line}
The proposed method (TFGNSCS) is constructed based on FGN\cite{YXian2018} framework, and extends two loss terms for building transfer feature generating networks. We implement the ablation study for comparing the proposed method with the base-line methods.  Therefore, the related base-line methods include FGN\cite{YXian2018}(the optimization function is $\min_{G}\max_{D}L=\min_{G}\max_{D}L_{WGAN}+\beta L_{CLS}$), TFGNSCS-1(the optimization function is $\min_{G}\max_{D}L=\min_{G}\max_{D}L_{WGAN}+\beta L_{CLS}+\gamma L_{TRA1}$) and TFGNSCS-2(the optimization function is $\min_{G}\max_{D}L=\min_{G}\max_{D}L_{WGAN}+\beta L_{CLS}+\eta L_{TRA2}$). In FGN, the model only considers of the classification loss ($L_{CLS}$) of seen classes based on the generative adversarial framework. In TFGNSCS-1, the model integrates the classification loss ($L_{TRA1}$) of unseen classes with FGN model based on semantic structure transfer. In TFGNSCS-2, the model combines the discriminator loss ($L_{TRA1}$) for unseen classes with FGN model. TFGNSCS model considers all of these factors for transferring and balancing the information between seen and unseen classes.

\begin{table}[!ht]
\small
\renewcommand{\arraystretch}{1.0}
\caption{Comparison of TFGNSCS method with the base-line methods for ZSL with semantic feature and ResNet visual feature.}
\label{table4}
\begin{center}
\newcommand{\tabincell}[2]{\begin{tabular}{@{}#1@{}}#2\end{tabular}}
\begin{tabular}{lp{1cm}p{1cm}p{1cm}p{1cm}}
\hline
\bfseries  &\bfseries ZSL &\bfseries  &\bfseries &\bfseries \\
\cline{2-5}
\bfseries Method &\bfseries AwA &\bfseries CUB &\bfseries SUN &\bfseries FLO  \\
\cline{2-5}
\hline \hline
FGN\cite{YXian2018}  & $68.2$   &$57.3$  & $60.8$   &$67.2$ \\
\hline
TFGNSCS-1  & $69.4$   &$59.5$ & $63.7$   &$67.8$  \\
\hline
TFGNSCS-2  & $68.9$   &$58.6$ & $62.9$   &$67.3$  \\
\hline\hline
TFGNSCS & $\textbf{70.5}$   &$\textbf{60.2}$ & $\textbf{64.8}$ & $\textbf{68.1}$   \\
\hline
\end{tabular}
\end{center}
\end{table}

\begin{table}[!ht]
\small
\renewcommand{\arraystretch}{1.0}
\caption{Comparison of TFGNSCS method with the base-line methods for GZSL with semantic feature and ResNet visual feature. tr=average per-class Top-1 accuracy (\%) on seen classes, ts=average per-class Top-1 accuracy (\%) on unseen classes, H=harmonic mean for GZSL are reported based on the same data configurations in the different datasets splits. }
\label{table5}
\begin{center}
\newcommand{\tabincell}[2]{\begin{tabular}{@{}#1@{}}#2\end{tabular}}
\begin{tabular}{cp{0.4cm}p{0.4cm}p{0.4cm}|p{0.4cm}p{0.4cm}p{0.4cm}|p{0.4cm}p{0.4cm}p{0.4cm}|p{0.4cm}p{0.4cm}p{0.4cm}}
\hline
\bfseries   &\multicolumn{12}{c}{GZSL} \\ \cline{2-13}
\bfseries&\bfseries AwA &\bfseries &\bfseries &\bfseries CUB &\bfseries &\bfseries &\bfseries SUN &\bfseries &\bfseries &\bfseries FLO &\bfseries &\bfseries \\
\cline{2-13}
\bfseries Method &\bfseries ts &\bfseries tr &\bfseries H &\bfseries ts &\bfseries tr &\bfseries H &\bfseries ts &\bfseries tr &\bfseries H &\bfseries ts &\bfseries tr &\bfseries H\\
\hline \hline
FGN\cite{YXian2018}  & $57.9$ & $61.4$ & $59.6$  &$43.7 $ & $57.7$ & $49.7$ &$42.6$ & $36.6$ & $39.4$ &$59.0$ & $73.8$ & $65.6$ \\
\hline
TFGNSCS-1  & $58.4$ & $\textbf{69.8}$ & $63.5$ & $46.7$ & $58.4$ & $51.8$ & $43.8$ & $\textbf{37.7}$ & $40.5$ & $60.1$& $77.6$ & $67.7$\\
\hline
TFGNSCS-2  & $58.2$ & $67.9$ & $62.6$ & $45.4$ & $59.6$ & $51.5$ & $43.0$ & $37.1$ & $39.8$ &$59.8$ &$77.2$ &$67.3$\\
\hline\hline
TFGNSCS & $\textbf{60.3}$ & $69.5$ & $\textbf{64.5}$ & $\textbf{47.6}$ & $\textbf{59.7}$ & $\textbf{53.0}$ & $\textbf{44.3}$ & $37.5$ & $\textbf{40.6}$ & $\textbf{61.1}$ & $\textbf{78.6}$ & $\textbf{68.7}$ \\
\hline
\end{tabular}
\end{center}
\end{table}

Tab.\ref{table4} shows the experimental results of the proposed method TFGNSCS and the base-line methods for ZSL. The performance TFGNSCS outperforms that of the other methods. The improvement of TFGNSCS is $2.3\%$ for AwA, $2.9\%$ for CUB, $4\%$ for SUN and $0.9\%$ for FLO at least. Tab.\ref{table5} demonstrates that the performance TFGNSCS outperforms that of the base-line methods for GZSL. $H$ of TFGNSCS respectively improves $1.0$ for AwA, $1.2$ for CUB, $0.1$ for SUN, and $1.0$ for FLO\emph{} at least. For ZSL and GZSL, the performance of TFGNSCS-1 is better than that of TFGNSCS-2 and FGN, while the performance of FGN is worse than that TFGNSCS-1 and TFGNSCS-2. $L_{TRA1}$ in TFGNSCS-1 is individually considered to ascend the performance of the model, and  $L_{TRA2}$ in TFGNSCS-2 is individually considered also to improve the performance of the model. However, $L_{TRA2}$ individually enhance the difference of the discriminator between seen and unseen classes, whereas transfer factor $L_{TRA1}$ can weaken these imbalance information between seen and unseen classes. Therefore, the combination of $L_{TRA1}$  and $L_{TRA2}$  in TFGNSCS can make $L_{TRA2}$ boost the transfer characteristic of $L_{TRA1}$ for improving the performance of ZSL and GZSL.

\subsection{Comparison with the different transfer methods}
\label{transfer-methods}
Transfer method is an key point for constructing generating network model. In this paper, we focus on the transfer method from the classifier model $P(\theta)$ of seen classes to the classifier model $Q(\theta)$ of unseen classes. The equation (\ref{model2}) shows the transformation relationship between $P(\theta)$ and $Q(\theta)$. Beside this way, while we incorporate image feature of unseen classes into a semantic class prototype graph, ZSL can be regard as an extended absorbing Markov chain process on this graph \cite{fu2018zero}. Therefore, a alternative transfer transformation function $T(P(\theta))$ is
\begin{align}
\label{model22}
\begin{aligned}
&Q(\theta)=T(P(\theta))=P(\theta)(I-W_{ss})^{-1}W_{su},
\end{aligned}
\end{align}
where $I\in \mathbb{R}^{K\times K}$. We use the transfer method to learn model, which can be expressed as TFGNSCS-alt. In Tab.\ref{table6} and Tab.\ref{table7}, we find that difference between transfer methods is slight for ZSL or GZSL. The main reason is that the different transfer methods can both adjust the imbalance information between seen and unseen classes for improving the discrimination of generative feature by adversarial learning.

\begin{table}[!ht]
\small
\renewcommand{\arraystretch}{1.0}
\caption{Comparison of TFGNSCS method with the alternative transfer method for ZSL with semantic feature and ResNet visual feature.}
\label{table6}
\begin{center}
\newcommand{\tabincell}[2]{\begin{tabular}{@{}#1@{}}#2\end{tabular}}
\begin{tabular}{lp{1cm}p{1cm}p{1cm}p{1cm}}
\hline
\bfseries  &\bfseries ZSL &\bfseries  &\bfseries &\bfseries \\
\cline{2-5}
\bfseries Method &\bfseries AwA &\bfseries CUB &\bfseries SUN &\bfseries FLO  \\
\cline{2-5}
\hline \hline
TFGNSCS-alt  & $69.7$   &$\textbf{60.8}$ & $63.4$   &$\textbf{68.3}$  \\
\hline\hline
TFGNSCS & $\textbf{70.5}$   &$60.2$ & $\textbf{64.8}$ & $68.1$   \\
\hline
\end{tabular}
\end{center}
\end{table}

\begin{table}[!ht]
\small
\renewcommand{\arraystretch}{1.0}
\caption{Comparison of TFGNSCS method with the alternative transfer method for GZSL with semantic feature and ResNet visual feature. tr=average per-class Top-1 accuracy (\%) on seen classes, ts=average per-class Top-1 accuracy (\%) on unseen classes, H=harmonic mean for GZSL are reported based on the same data configurations in the different datasets splits. }
\label{table7}
\begin{center}
\newcommand{\tabincell}[2]{\begin{tabular}{@{}#1@{}}#2\end{tabular}}
\begin{tabular}{cp{0.4cm}p{0.4cm}p{0.4cm}|p{0.4cm}p{0.4cm}p{0.4cm}|p{0.4cm}p{0.4cm}p{0.4cm}|p{0.4cm}p{0.4cm}p{0.4cm}}
\hline
\bfseries   &\multicolumn{12}{c}{GZSL} \\ \cline{2-13}
\bfseries&\bfseries AwA &\bfseries &\bfseries &\bfseries CUB &\bfseries &\bfseries &\bfseries SUN &\bfseries &\bfseries &\bfseries FLO &\bfseries &\bfseries \\
\cline{2-13}
\bfseries Method &\bfseries ts &\bfseries tr &\bfseries H &\bfseries ts &\bfseries tr &\bfseries H &\bfseries ts &\bfseries tr &\bfseries H &\bfseries ts &\bfseries tr &\bfseries H\\
\hline \hline
TFGNSCS-alter  & $\textbf{60.5}$ & $68.9$ & $64.4$ & $46.8$ & $\textbf{59.9}$ & $52.5$ & $43.9$ & $\textbf{37.8}$ & $\textbf{40.6}$ &$60.7$ &$\textbf{78.8}$ &$68.5$\\
\hline\hline
TFGNSCS & $60.3$ & $\textbf{69.5}$ & $\textbf{64.5}$ & $\textbf{47.6}$ & $59.7$ & $\textbf{53.0}$ & $\textbf{44.3}$ & $37.5$ & $\textbf{40.6}$ & $\textbf{61.1}$ & $78.6$ & $\textbf{68.7}$ \\
\hline
\end{tabular}
\end{center}
\end{table}

\subsection{Parameter analysis}
In TFGNSCS, the number of generative features directly impacts on the performance of ZSL or GZSL. Therefore we select the number of generative features from $1,2,6,10,30,50,100,200,300$ to construct the different classification model for comparing TFGNSCS with the base-line methods. Figure \ref{fig2} shows TFGNSCS outperforms the base-line methods. Especialy there is the significant boost of classification accuracy with number increasing from 1 to 50 of generative features for unseen classes, e.g. $42.1$ to $67.7$ on FLO and $33.3$ to $57.0$ on CUB in TFGNSCS. Figure \ref{fig3} also demonstrates the significant boost of harmonic mean with number increasing from 1 to 50 of generative features for unseen classes, e.g. $48.1$ to $67.7$ on FLO and $32.1$ to $48.6$ on CUB in TFGNSCS. It shows that TFGNSCS has the better adaptability than other methods.

\begin{figure*}[ht]
 \begin{center}
\includegraphics[width=1\linewidth]{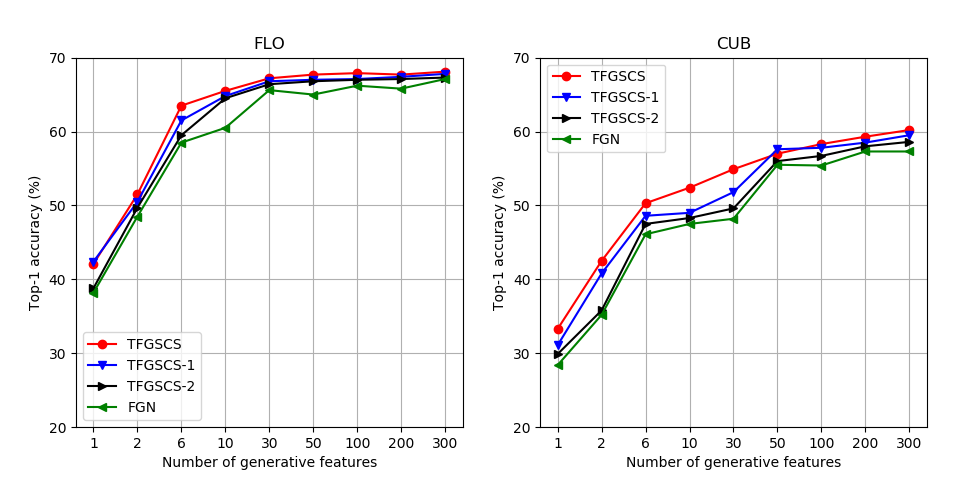}
\vspace{-0.2in}
 \caption{Impact of the generative features number on unseen class accuracy for zero-shot learning on FLO and CUB.}
  \label{fig2}
  \end{center}
 \end{figure*}

\begin{figure*}[ht]
\begin{center}
\includegraphics[width=1\linewidth]{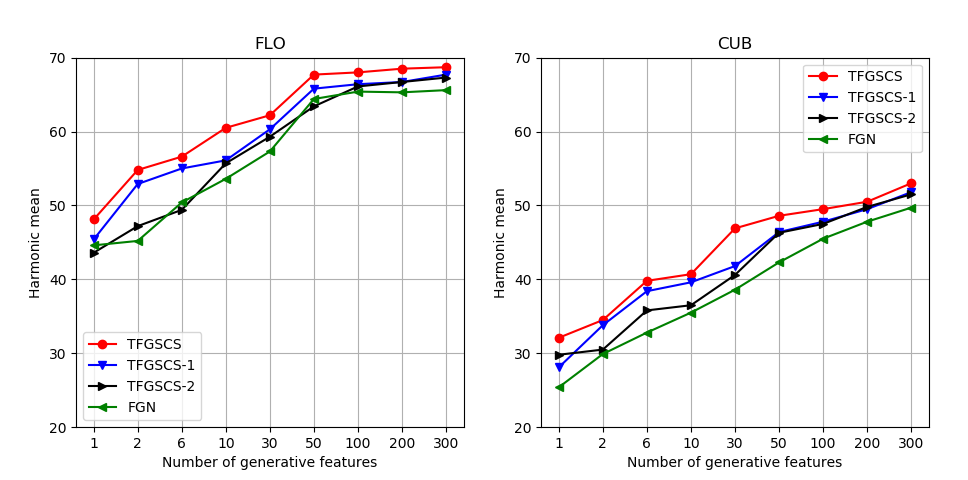}
\end{center}
\vspace{-0.2in}
 \caption{Impact of the generative features number on harmonic mean for generalization zero-shot learning on FLO and CUB.}
  \label{fig3}
 \end{figure*}

\subsection{Experimental results analysis}
\label{analysis}
In experiments, we compare the proposed method with eight methods, which include five kinds of state-of-the-art methods (DVN \cite{ZHANG201943},HM\cite{ZHANG201912},VCL\cite{wan019},TVN\cite{8464092} and FGN\cite{YXian2018} in section \ref{State-of-the-arts}), three kinds of base-line methods(FGN\cite{YXian2018},TFGNSCS-1 and TFGNSCS-2 in section \ref{base-line}) and a alternative transfer method (TFGNSCS-alt in section \ref{transfer-methods}). These methods can construct the related model to bridge the gaps between visual and semantic information for ZSL or GZSL. In contrast to these methods, the proposed method focuses on mining the transfer information in generation adversarial framework for the discriminative synthetic feature. From these experiment, we have the following observations.

\begin{itemize}
\item The performance of TFGNSCS is better than that of five kinds of state-of-the-art methods (DVN,HM,VCL,TVN and FGN in section \ref{State-of-the-arts}) for ZSL, and the performance of TFGNSCS outperforms that of these methods for GZSL. This situation of the main reason is that TFGNSCS try to balance the difference between seen and unseen classes by transfer losses. Therefore, in ZSL setting, the classification accuracy of unseen classes is higher than other methods on four datasets, moreover, the betterment is noticeable for harmonic mean in GZSL setting. Especially, harmonic mean of FGN and TFGNSCS significantly exceeds that of other state-of-the-art methods for GZSL in the different datasets.
\item The performance improvement of TFGNSCS is different in three base-line approaches (FGN,TFGNSCS-1,TFGNSCS-2) for ZSL or GZSL. The advance of TFGNSCS can be found for ZSL in four datasets, while the better improvement can be demonstrated for GZSL in all datasets. In there,the outstanding betterment is harmonic mean of GZSL in AwA and FLO. It shows that transfer method can enhance the discrimination of the generative features by transfer losses in adversarial networks, and further validates that semantic structure transfer is effective for constructing the learning model in ZSL or GZSL.
\item The different transfer methods have the similar performance for ZSL and GZSL in all datasets. In this paper, two kinds of transfer method both is built based on the semantic classes of graph, which can represent the distribution structure of classes. The difference of these transfer methods is on the different way of the structure propagation. The structure propagation of that equation (\ref{model2}) in TFGNSCS is formed by drawing support from the relationship of the sharing parts in respective classification model on the seen or unseen classes, while the structure propagation of that equation (\ref{model22}) in TFGNSCS-alt is constructed based on an extended absorbing Markov chain process. In the learning process of the adversarial networks, this difference of the structure propagation is trivial for ZSL and GZSL.
\item The two loss parts (the unseen classes classification loss $L_{TRA1}$ and the generated features discrimination loss $L_{TRA2}$ ) of transfer loss have the diverse effect to improve the performance of ZSL and GZSL. The unseen classes classification loss $L_{TRA1}$ can  boost the performance of ZSL and GZSL, while the generated features discrimination loss $L_{TRA2}$  play a fewer role for this melioration. However, the integration of these loss can further ameliorate the performance of ZSL and GZSL. The loss $L_{TRA2}$ is a assistant method for ascending the performance of TFGNSCS, and it's role depends on the quality of the generative feature by $L_{TRA1}$.
\item The number of the generative features influences the performance of ZSL and GZSL. The increasing number of the generative features makes the performance improve for ZSL, and this situation is more obvious for GZSL. It shows that the proposed method TFGNSCS can synthesize the more discriminative feature at the same number because of the transfer loss contribution for model construction.

\end{itemize}

\section{Conclusion}
We have proposed transfer feature generating networks with semantic classes structure (TFGNSCS) method to address imbalance between seen and unseen classes in ZSL and GZSL. TFGNSCS can not only adapt the semantic structure relationship between seen and unseen classes to a uniform generative features framework, but also model the difference of generating features by balancing
transfer information between seen and unseen classes in networks. Furthermore, TFGNSCS can combine a Wasserstein generative adversarial network with classification loss and transfer loss to generate enough CNN feature for improving ZSL and GZSL. At last, the optimization learning of the TFGNSCS can obtain both the transfer feature generating networks and the more discriminative features. For evaluating the proposed TFGNSCS, we carry out the comparison experiments about the state of the art methods, the baseline methods, the other transfer method and parameter analysis on AwA, CUB, SUN and FLO datasets. Experiment results demonstrate the TFGNSCS gets the promising results in ZSL and GZSL.

\section{Acknowledgements}
The authors would like to thank the anonymous reviewers for their insightful comments that help improve the quality of this paper. Especially, The authors thank to Dr. Yongqin Xian from MPI for Informatics, who provided the data source to us. This work was supported by NSFC (Program No.61771386,Program No.61671376 and Program No.61671374), Natural Science Basic Research Plan in Shaanxi Province of China (Program No.2017JZ020).

\section*{References}

\bibliography{mybibfile}

\end{document}